\documentclass{article}       

\usepackage[backend=biber,style=numeric]{biblatex}
\usepackage{amsmath}
\usepackage{amsthm}
\usepackage{amsfonts}
\usepackage{mathtools}
\usepackage{amssymb}

\usepackage[utf8]{inputenc}
\usepackage{todonotes}
\usepackage{url}

\usepackage{xspace}
\newcommand{\mathsymbol}[2]{\newcommand{#1}{\ensuremath{\mathit{#2}}\xspace}}
\newcommand{\textmacro}[2]{\mathsymbol{#1}{\text{#2}}}
\mathsymbol{\hyperband}{\text{\sc Hyperband}}

\mathsymbol{\dataspace}{\mathcal{D}}
\mathsymbol{\searchspace}{\mathcal{P}}
\mathsymbol{\instancespace}{\mathcal{X}}
\mathsymbol{\labelspace}{\mathcal{Y}}
\mathsymbol{\risk}{\mathcal{R}}
\mathsymbol{\performance}{\phi}
\mathsymbol{\void}{\bot}

\textmacro{\blackbox}{black-box}

\textmacro{\automl}{AutoML}
\textmacro{\naml}{Naive \automl}
\textmacro{\mlplan}{ML-Plan}
\textmacro{\autoweka}{Auto-WEKA}
\textmacro{\autosklearn}{auto-sklearn}
\textmacro{\tpot}{TPOT}
\textmacro{\recipe}{RECIPE}
\textmacro{\exdef}{``Ex-def''}

\usepackage{booktabs}

\begin{document}




\author{Felix Mohr, Marcel Wever}
\title{Naive Automated Machine Learning - A Late Baseline for \automl}
\maketitle

\begin{abstract}
Automated Machine Learning (\automl) is the problem of automatically finding the pipeline with the best generalization performance on some given dataset.
\automl has received enormous attention in the last decade and has been addressed with sophisticated \blackbox optimization techniques such as Bayesian Optimization, Grammar-Based Genetic Algorithms, and tree search algorithms.
In contrast to those approaches, we present \naml, a very simple solution to \automl that exploits important meta-knowledge about machine learning problems and makes simplifying, yet, effective assumptions to quickly come to high-quality solutions.
While \naml can be considered a baseline for the highly sophisticated \blackbox solvers, we empirically show that those solvers are not able to outperform \naml; sometimes the contrary is true.
On the other hand, \naml comes with strong advantages such as interpretability and flexibility and poses a strong challenge to current tools.
\end{abstract}

\section{Introduction}
Automated machine learning (\automl) is the problem of automatically finding the data transformation and learning algorithms with the best generalization performance on a given dataset.
The combination of such algorithms is typically called machine learning \emph{pipeline}, because several algorithms for data manipulation and analysis are concatenated sequentially.
To optimize such a machine learning pipeline, important decisions do not only include the learning algorithm itself but also its parameters as well as a definition of which features should be used by the learner.
The latter decision is often implicitly resolved by choosing and configuring feature pre-processing algorithms.

Surprisingly, all common \automl approaches solve the problem with a general-purpose \blackbox optimization technique without explicitly using knowledge about the \automl problem \cite{autoweka,feurer2015efficient}.
The main argument behind this approach and the alleged superiority of \automl tools over human data scientists is that a lot of decisions for the pipeline need to be made \emph{simultaneously}, which cannot be easily achieved in a sequential optimization process a human expert would pursue.
The surprising part is that imitating a human data scientist seems a natural baseline, but never have the outcomes of such an imitation process been used as a baseline for the complex tools presented in the community.
In \cite{autoweka}, at least the attempt is made to the degree of running a cross-validation over different learners, but no further steps a data scientist would usually conduct are considered.

In this paper, we make up for this natural \automl baseline, which imitates the process a human expert might go through to find the best pipeline.
Instead of crawling an enormous search space with a \blackbox optimizer that iteratively evaluates full pipeline candidates to finally pick the best one, the idea is to conduct an analytical process and \emph{derive} the appropriate pipeline from it.
In this, we implicitly assume a form of independence of the optimization decisions and hence act a bit \emph{naive} w.r.t. potential interactions between them; this is why we call the approach \naml.
In this sense, we complement the \exdef baseline \cite{autoweka} by what a somewhat more experienced data scientist would maybe do.

Our experimental evaluation evidences the strength of this simple approach and strongly jeopardizes current \automl tools.
\naml is highly competitive with the state of the art and is sometimes even significantly better.
In fact, the experiments convincingly show that, other than reported in \cite{autoweka}, even the simple baseline of only trying out once each base learner with its default configuration is often \emph{not} outperformed by the sophisticated optimizers.
This is a somewhat puzzling result, because the justifications to use the current \automl tools seem to disappear.
\naml achieves comparable or better results, is much more simple to understand, more flexible to extend, and easy to implement.
Taking all these observations together, \naml imposes a strong challenge to currently advocated approaches.

At this point, we shall stress that the main goal of this paper is \emph{not} to present yet another tool for \automl but to propose a different \emph{view} on how automated machine learning could be done.
Recent work has revealed that several aspects of the \blackbox character of current systems impede experts from trusting their output, and that there is a need for more interaction between the tool and the expert\cite{wang2019human,crisan2021fits,drozdal2020trust,wang2021autods}.
Since \naml effectively \emph{imitates} a data scientist, it makes an important step into this direction by adopting an optimization process that is much more comprehensive for an expert who interacts with it.
Even if \naml was less competitive than it is, we would still consider this perspective an important contribution of the approach by itself.
It paves the way for completely new approaches that are inconceivable with the current \blackbox techniques.

\section{Problem Definition}
\label{sec:problem}
Even though the vision of \automl is much broader, the \automl problem in the narrower sense as addressed by most \automl approaches is to automatically \emph{compose} and \emph{parametrize} machine learning algorithms to maximize a given metric such as predictive accuracy.
The available algorithms are typically related either to preprocessing (feature selection, transformation, imputation, etc.) or to the core functionality (classification, regression, ranking, etc.).

In this paper, we are focused on \automl for supervised learning.
Formally, in the supervised learning context, we assume some \emph{instance space} $\instancespace \subseteq \mathbb{R}^d$ and a \emph{label space} \labelspace.
A \emph{dataset} $D \subset \{(x,y)~|~x\in \instancespace, y \in \labelspace\}$ is a \emph{finite} relation between the instance space and the label space, and we denote as \dataspace the set of all possible datasets.
We consider two types of operations over instance and label spaces:
\begin{enumerate}
    \item \emph{Transformers.}
    A transformer is a function $t: \instancespace_A \rightarrow \instancespace_B$, converting an instance $x$ of instance space $\instancespace_A$ into an instance of another instance space $\instancespace_B$.
    \item \emph{Predictors.}
    A predictor is a function $p: \instancespace_p \rightarrow \labelspace$, assigning an instance of its instance space $\instancespace_p$ a label of the original label space \labelspace.
\end{enumerate}
In this paper, a \emph{pipeline} $P = t_1 \circ .. \circ t_k \circ p$ is a functional concatenation in which $t_i: \instancespace_{i-1} \rightarrow \instancespace_i$ are transformers with $\instancespace_0 = \instancespace$ being the original instance space, and $p: \instancespace_k \rightarrow \labelspace$ is a predictor.
Hence, a pipeline is a function $P: \instancespace \rightarrow \labelspace$ that assigns a label to each object of the instance space.
We denote as \searchspace the space of all pipelines of this kind.
In general, the first part of a pipeline could not only be a sequence but also a \emph{transformation tree} with several parallel transformations that are then merged \cite{olson2016tpot}, but we do not consider such structures in this paper since they are not necessary for our key argument.
An extension to such tree-shaped pipelines is canonical future work.

In addition to the sequential structure, many \automl approaches restrict the search space still a bit further.
First, often a particular best order in which different \emph{types} of transformation algorithms should be applied is assumed.
For example, we assume that feature selection should be conducted after feature scaling.
So \searchspace will only contain pipelines compatible with this order.
Second, the optimal pipeline uses at most one transformation algorithm of each type.
For example, no optimal pipeline will concatenate two feature scalers.
These assumptions allow us to express \emph{every} element of \searchspace as a concatenation of $k+1$ functions, where $k$ is the number of considered transformation algorithm \emph{types}, e.g., feature scalers, feature selectors, etc.
If a pipeline does not adopt an algorithm of one of those types, say the $i$-th type, then $t_i$ will simply be the identity function.

The theoretical goal in supervised machine learning would be to find
\begin{equation}\label{eq:cs}
P^*~ \in \underset{P \in \searchspace}{\operatorname{argmin}} ~\risk \big(P \big) \, ,
\end{equation}
with the \emph{risk} or expected loss of the pipeline $P$ given by
\begin{equation}\label{eq:risk}
\risk(P) = \int\limits_{\instancespace,\labelspace}{loss(y, P(x))} \, d \mathbb{P}(x,y) \, .
\end{equation}
Here, $loss(y,P(x)) \in \mathbb{R}$ is the penalty for predicting $P(x)$ for instance $x \in \instancespace$ when the true label is $y \in \labelspace$, and $\mathbb{P}$ is a joint probability measure on $\instancespace \times \labelspace$ from which the available dataset $D$ has been generated.

In practice, (\ref{eq:risk}) cannot be evaluated as the data-generating process $\mathbb{P}$ is assumed to exist but is obviously not known to the learner.
Hence, the true performance \risk is replaced by some performance estimator $\performance: \dataspace \times \searchspace \rightarrow \mathbb{R}$ that estimates the performance of a candidate pipeline based on some validation data.
A typical metric used for this evaluation is the average loss on the validation fold, but, based on the exact problem type, other metrics such as least squares, AUC/ROC, F1 measure, or others are conceivable.

Consequently, a supervised \automl problem \emph{instance} is defined by a dataset $D\in \dataspace$, a search space \searchspace of pipelines, and a performance estimation metric $\performance: \dataspace \times \searchspace \rightarrow \mathbb{R}$ for solutions.
An \automl \emph{solver} $\mathcal{A}: \dataspace \rightarrow \searchspace$ is a function that creates a pipeline given some training set $D_{train} \subset D$.
The performance of $\mathcal{A}$ is given by

\begin{equation}
    \mathbb{E}\left[ ~\performance \big(D_{test}, \mathcal{A}(D_{train}) \big) \right]\, ,
\end{equation}
where the expectation is taken with respect to the possible (disjoint) splits of $D$ into $D_{train}$ and $D_{test}$.
In practice, this score is typically computed taking a series of random binary splits of $D$ and averaging over the observed scores.
Naturally, the goal of any \automl solver is to optimize this metric, and we assume that $\mathcal{A}$ has access to $\phi$ (but not to $D_{test}$) in order to evaluate candidates with respect to the objective function.


\section{Related Work}
Even though the foundation of \automl is often attributed to the proposal of \autoweka, there have been some works on the topic before.
These initial approaches have been based on tree search.
Here, a large search tree is given in which each path encodes a machine learning pipeline.
There are mainly three approaches in this direction, differing in the way how the search space is defined and how the search process is guided.
The very first approach we are aware of was designed for the configuration of RapidMiner modules based on hierarchical planning \cite{kietz2009towards,kietz2012designing} most notably MetaMiner \cite{nguyen2011meta,nguyen2012experimental,nguyen2014using}.
With ML-Plan \cite{mohr2018ml}, the idea of HTN-based graph definitions was later combined with a best-first search using random roll-outs to obtain node quality estimates.
Similarly, \cite{DBLP:conf/ijcai/RakotoarisonSS19} introduced \automl based on Monte-Carlo Tree Search, which is closely related to ML-Plan.
However, the authors of \cite{DBLP:conf/ijcai/RakotoarisonSS19} do not discuss the layout of the search tree, which is a crucial detail, because it is the main channel to inject knowledge into the search problem.

Another line of research based on Bayesian Optimization was initialized with the advent of \autoweka \cite{autoweka,kotthoff2017auto}.
Like \naml, \autoweka assumes a fixed structure of the pipeline, admitting a feature selection step, and a predictor.
The decisions are encoded into a large vector that is then optimized using SMAC \cite{hutter2011sequential}.
\autoweka optimizes pipelines with algorithms of the Java data analysis library WEKA \cite{hall2009weka}.
For the Python framework scikit-learn \cite{pedregosa2011scikit}, the same technique was adopted by \autosklearn \cite{feurer2015efficient}.
In the original version, \autosklearn added a data transformation step to the pipeline; meanwhile the tool has been extended to support some more pre-processing functionalities in the pipeline.
Besides, \autosklearn features warm-starting and ensembling.
The main difference between these approaches and tree search is that tree search successively \emph{creates} solution candidates as paths of a tree instead of obtaining them from an \emph{acquisition function} as done by \autoweka and \autosklearn.

The idea of warm-starting introduced by \autosklearn was also examined in specific works based on recommendations.
Approaches here include specifically collaborative filtering like OBOE \cite{yang2019oboe} and probabilistic matrix factorization \cite{fusi2018probabilistic}.
These approaches not necessarily require but are specifically designed for cases in which a database of past experiences on other datasets is available.



Another interesting line of research is the application of evolutionary algorithms.
One of these approaches is TPOT \cite{olson2016tpot}.
In contrast to the above approaches, \tpot allows not just one pre-processing step but an arbitrary number of feature extraction techniques at the same time.
TPOT adopts a genetic algorithm to find good pipelines and adopts the scikit-learn framework to evaluate candidates.
Another approach is \recipe \cite{de2017recipe}, which uses a grammar-based evolutionary approach to evolve pipeline construction.
In this, it is similar to the tree search based approaches.
Another genetic approach, focused on the construction of stacking ensembles, was presented in \cite{chen2018autostacker}.

A recent line of research adopts a type of \blackbox optimization relying on the framework of multipliers (ADMM) \cite{boyd2011distributed}.
The main idea here is to decompose the optimization problem into two sub-problems for different variable types, taking into account that algorithm selection variables are Boolean while most parameter variables are continuous.
This approach was first presented in \cite{liu2020admm}.

All of the above approaches can be considered almost entirely \blackbox optimization approaches.
Some approaches like ML-Plan indirectly allow the incorporation of domain knowledge through the search tree definition, and almost all tools make some algorithmic adjustments to at least somewhat inject knowledge to the overall process (like warm-starting, ensemble building, etc.).

Notably, \autoweka is the only tool that has been evaluated against an (automated) data scientist baseline.
In the \autoweka publications, this baseline is called \exdef, and is simply a for-loop that iterates once over each learner, assesses their performance in some cross-validation, and chooses the best one.
This is identical to our probing stage as detailed in Sec. \ref{sec:stage:probing}.
We are not aware of another work that ever came back to a simulated data scientist baseline.

However, we have two concerns against the \exdef baseline.
First, in our experiments, we could not reproduce the weak performance of this automated amateur data scientist and found even this strategy to perform well quite often.
But even if this baseline is outperformed, we argue that the goal of beating this performance is not ambitious enough.
An expert data scientist would hardly ever conduct like \exdef and stop after this initial analysis but conduct a series of further operations.
In \cite{autoweka}, a second baseline includes a grid search, but many data scientists would probably rather focus on data transformation instead of parameter tuning.
In this paper, we implement such a stronger baseline and show that the state-of-the-art approaches hardly beat \emph{this} baseline.

\section{\naml}

\subsection{Naivety Assumptions}
\label{sec:assumptions}
\naml makes, among others, the assumption that the optimal pipeline is the one that is locally best for each of its transformers and the final predictor.
In other words, taking into account pipelines with (up to) $k$ transformers and a predictor, we assume that for all datasets $D$ and all $1\leq i\leq k+1$
$$c_i^* \in \arg \min_{c_i} \performance(D, c_1\circ..\circ c_{k+1})$$
is \emph{invariant} to the choices of $c_1,..c_{i-1},c_{i+1},..,c_{k+1}$, which are supposed to be fixed in the above equation.
Note that we here use the letter $c$ instead of $t$ for transformers or $p$ for the predictor, because $c$ may be any of the two types.

We dub the approach \naml, because there is a direct link to the assumption made by the Naive Bayes classifier.
Consider \searchspace an urn and denote as $Y$ the event to observe an optimal pipeline in the urn.
Then
$$\mathbb{P}(Y~|~c_1,..,c_{k+1}) \propto \mathbb{P}(c_1,..,c_{k+1}~|~Y)\mathbb{P}(Y) \overset{naive}{=}\mathbb{P}(c_i~|~Y)\prod_{j=1,j\neq i}^{k+1}\mathbb{P}(c_j~|~Y)\mathbb{P}(Y),$$
in which we consider $c_j$ to be fixed components for $j \neq i$, and only $c_i$ being subject to optimization.
Applying Bayes theorem again to $\mathbb{P}(c_i~|~Y)$ and observing that the remaining product is a constant regardless the choices of $c_{i\neq j}$, it gets clear that the optimal solution is the one that maximizes the probability of being locally optimal, and that this choice is \emph{independent} of the choice of the other components.

The typical approach to optimize the $c_i$ is not to directly construct those functions but to adopt parametrized model building processes that create these functions.
For example, $c_1$ could be a projection obtained by determining some features which we want to stay with, or $c_{k+1}$ could be a trained neural network.
These induction processes for the components can be described by an algorithm $a_i$ and a parametrization $\theta_i$ of the algorithm.
The component $c_i$ is obtained by running $a_i$ under parameters $\theta_i$ with some training data.
So to optimize $c_i$, we need to optimally choose $a_i$ and $\theta_i$.

Within this regime, \naml makes the additional assumption that even each component $c_i$ can be optimized by local optimization techniques.
More precisely, it is assumed that the algorithm that yields the best component when using the default parametrization is also the algorithm that yields the best component if all algorithms are run with the best parametrization possible.

Observe that this assumption is less far-fetched than one might expect.
Our preliminary experiments showed that the tuning of parameters has often no or only a slim improvement over the performance achieved with the default configuration.
Literature reports some exceptions, but these are mostly in the area of neural networks, in which the network architecture is considered as a parameter and clearly has a substantial impact on the performance.
However, for most learners, the variance of the variable describing the \emph{improvement} of a configuration over the performance with default configuration is rather low.
And even if we consider some parameters so important that changing them effectively completely changes the algorithm behavior, consider these instantiations as different algorithms.
For example, we could simply treat support vector machines with different kernels and different (orders of) complexity constants as different algorithms.
In the large majority of algorithms, this practice does not yield an explosion in the algorithm space.
In fact, the only exception we can think of is indeed neural networks.


\subsection{The \naml Stage Scheme}
The idea of \naml is to conduct a series of optimization \emph{stages}.
Each stage optimizes one aspect of the pipeline.
For example, one stage chooses a predictor, and another optimizes its parameters.
The order of stages does not need to coincide with the order of the elements in the pipeline, and we also admit that there are several stages for the \emph{same} aspect.
In a strictly naive setting, the latter one is not necessary, but it can make sense to adjust an initial decision later when other decisions have been made.

Each stage produces a \emph{candidate pool} and \emph{may} be based on the candidate pool (and other results) of the preceding stages.
In \naml, a \emph{candidate} is simply an encoding that contains all the information necessary to build and evaluate the corresponding pipeline.
For example, in our concrete prototype presented in Sec. \ref{sec:prototype}, a candidate is encoded by a tuple $\mathit{(scale, F, a, \theta)}$, where $\mathit{scale}$ encodes the feature transformations (applied to all columns), $F$ is a \emph{set} of feature indices to consider, $a$ is the prediction algorithm, and $\theta$ the parametrization for $a$.

\begin{figure}[t]
    \centering
    \includegraphics[width=.65\textwidth]{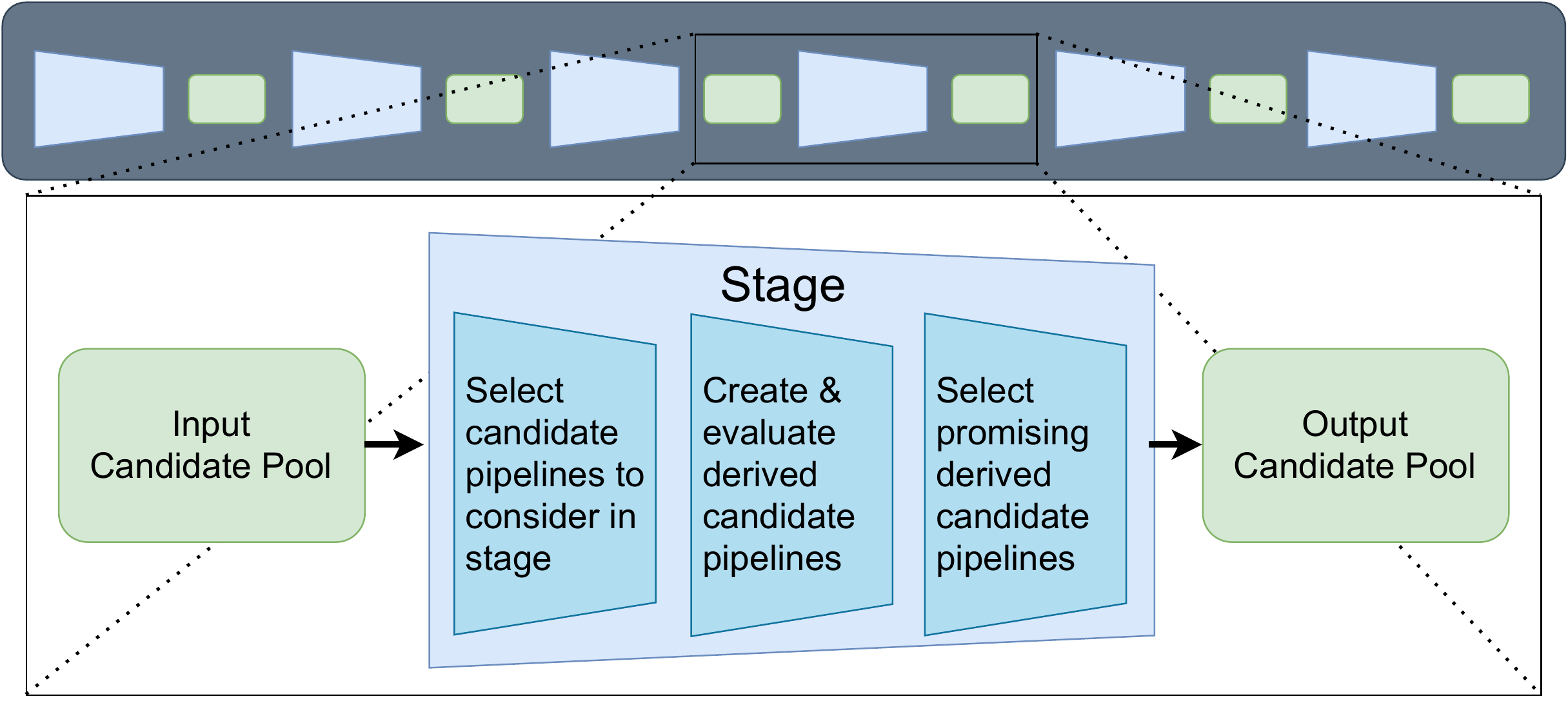}
    \caption{Generic schema of a single stage, taking a candidate pool as input and eventually returning an augmented candidate pool.}
    \label{fig:stageschema}
\end{figure}

While every stage is entirely free in the way how it modifies the candidate pool, there is a general pattern underlying the behavior of each stage.
We illustrate this pattern in Fig. \ref{fig:stageschema}, in which candidate pools are green and stages are blue.
The general pattern is that a stage creates and evaluates a new set of candidates, potentially taking into account existing candidates, and finally may also erase some candidates produced by itself or by previous stages.
Candidates from the incoming candidate pool might be ignored, e.g. if we just want to try some entirely new type of pipelines.
But we can also consider the \emph{whole} incoming pool, e.g. when combining every existing candidate with a particular concept for feature selection.
Often, a stage will \emph{sort} the candidates of the incoming pool by their performance, and consider them one by one.
For example, in a parameter tuning stage, we might tune the parameters of the algorithms of the incoming candidates, one by one in the order of their previous performance.
At the end of the stage, we might eliminate some candidates.
For example, a parameter tuning stage may generate a large set of new candidates, but it is reasonable to stay only with a small portion of them.
It can also make sense to eliminate candidates that came with the input pool.
For example, if we have an initial stage that only considers learners with configurations that enable a very fast runtime, we might want to replace them later when other parametrized versions of them have been tried.

With respect to time limitations, \naml seems to be more flexible than the \blackbox approaches, because time can be allocated with respect to certain optimization steps.
That is, besides a global timeout, \naml of course also allows specifying timeouts \emph{per stage} or even for steps \emph{inside} of a stage.
For example, if we have a stage to tune the learner hyper-parameters, it usually makes sense to limit the time consumed in this stage and, on top of that, to limit the time used to optimize parameters \emph{per} candidate.
The overall runtime for the stage could be 10 minutes, with a time budget of 2 minutes for each learner that is optimized.

The output of \naml can be obtained by taking the candidate with the best solution observed either in \emph{any} stage or in the \emph{last} stage.
The default output is simply the best candidate that has been observed during the process in \emph{any} stage.
However, we could also introduce a (final) stage with the sole purpose of \emph{validation and model selection}.
For this purpose, we would hold back data from the beginning that is not used in all but the last stage.
In this case, we would of course use the best candidate of the last pool only (these scores are not comparable to those of the other stages then anyway).

The candidate pools enable a certain degree of communication between the stages, which might seem a bit contradictory to a strictly independent optimization.
However, some optimization steps are just logically dependent.
For example, it does not make sense to optimize hyperparameters of the predictor before we have chosen a predictor (unless we optimize for all possible candidates).
But even if such logical dependencies do not exist, we argue that our goal here is not to implement a strictly naive approach but to rather mimic an expert data scientist's optimization workflow.
While the expert's process is still largely naive in the way how it implicitly prunes large spaces of the search space, it will also employ \emph{some} interaction between the stages.
For example, we could not only select the features to stick with for the case of the original data but also for features that have been scaled before, and where the scaling was determined in a previous stage.
This interaction is still fairly reasonable but far away from simultaneous optimization of feature scaling and feature selection and can still be considered ``largely'' naive.

\subsection{A Prototype for a Stage Scheme}
\label{sec:prototype}
In this paper, we work with a specific \emph{exemplary} stage scheme.
The purpose of this scheme is only to illustrate the scheme idea with a concrete example and to provide a means of evaluating the idea of \naml.
However, this scheme is \emph{not} what we refer to as \naml but only \emph{one prototypical instantiation} of the \naml approach.
We certainly do not claim this prototype to be the last answer in \naml; better schemes can be found in future work.

\begin{figure}[t]
    \centering
    \includegraphics[width=\textwidth]{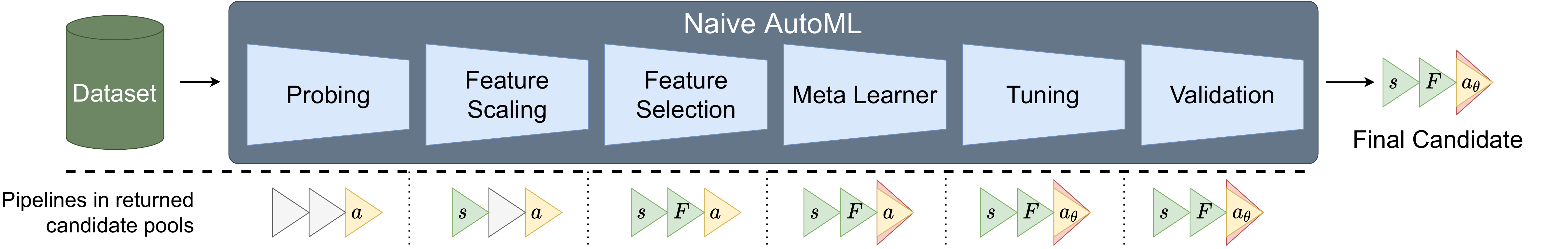}
    \caption{Visualization of the stages of the \naml machine learning pipeline (top), taking a dataset as an input and returning a final candidate. The lower part of the figure illustrates the potential shapes of machine learning pipelines contained in the candidate pool after the execution of the respective stage.}
    \label{fig:stageschema:prototype}
\end{figure}

Our prototype stage scheme is sketched in Fig. \ref{fig:stageschema:prototype}.
The first stage simply cross-validates every learning algorithm once without setting any of its hyper-parameters.
This stage corresponds to the \exdef baseline used in \cite{autoweka}.
The second stage simply pairs all feature scalers with all learners and observes their performance.
The third stage is independent of the first two stages and adopts filtering techniques to identify a subset of the features that are expected to bring the best performance (on average).
The fourth stage combines all previous candidates with each homogeneous meta learner, and the fifth stage tunes the \emph{learner} parameters of the incoming candidates (with a given time budget).
In the final stage, the most promising candidates are evaluated against a validation dataset.
Below, we discuss each of these stages in detail.

This stage scheme produces pipelines with up to three components.
A candidate is encoded by a tuple $\mathit{(s, F, a, \theta)}$, where $\mathit{s}$ encodes the feature transformations (applied to all columns), $F$ is a \emph{set} of feature indices to consider, $a$ is the prediction algorithm, and $\theta$ the parametrization for $a$.
The reason why $s$ is a function but $F$ is a set is simply that the feature selection is a projection that is entirely described by the features it shall retrain, but the scaling implies a functional transformation of data that cannot be captured so easily.
We will use the symbol \void to indicate that a choice has been left blank.

Despite being only an example scheme, one advantage of \naml over \blackbox optimization that already becomes clear here is that it directly generates important insights that can significantly support the data scientist working with it.
For example, even such a simple question as ``what is the potential of feature selection on the given data?'' cannot be directly answered by the existing \blackbox approaches.
In our scheme, the filtering stage discussed in Sec. \ref{sec:filteringstage} is a very good basis to give an initial answer to this question.
More complex stage schemes, e.g. including \emph{wrapping}, can answer such questions even in much more detail.
In this sense, \naml presents itself as more amenable to the growing demand for meaningful \emph{interaction} between the tool and the human \cite{wang2019human,crisan2021fits,drozdal2020trust,wang2021autods} compared to the currently adopted \blackbox approaches.

In the following sub-sections, we will assume that candidates are evaluated with some standard evaluation scheme.
A typical choice to evaluate candidates is to use standard cross validation techniques.
For example, in the evaluation in Sec. \ref{sec:evaluation}, we conduct a Monte-Carlo Cross Validation (MCCV) with a train fold size of 70\% and 5 repetitions.
This means that we build 5 random splits of 70\% training data and 30\% validation data each, train and test over the 5 folds using the desired metric, and then average these observations to get to a score.
However, \naml is not committed to a particular type of scoring function and could, for example, also be run with a 10-fold cross validation.
We hence just shall assume that every pipeline has some \emph{score}, and we leave it to the concrete implementation to implement one or another method.

\subsubsection{The Probing Stage}
\label{sec:stage:probing}
This stage corresponds to the \exdef suggested in the evaluation of \cite{autoweka}.
Formally, we create and evaluate \emph{all} candidates of the form $(\void, \void, a, \void)$, where $a$ is one of the available \emph{base} prediction algorithms.
Here, we do not consider ensemble learners like Boosting or Bagging but only those that are already implemented with a specific base learner, such as Random Forests.

\subsubsection{The Feature-Scaling Stage}
    In this step, we examine the benefit of different feature scaling operations for the base learners.
    To this end, two or three cheap distance-sensitive pilot classifiers like kNN or SVMs are used to assess the impact of different feature scaling techniques such as (mean)-normalization, standardization (mean 0 and std 1), or quantile-based re-scaling.
    In addition, it can make sense to add as a pilot the one or two best candidates resulting from the probing stage.
    Formally, for each scaler $s$ and each pilot algorithm $a_p$, we evaluate the candidate $(s, \void, a_p,\void)$.
    
    For each scaling technique $s$, if at least one pilot classifier improves upon its result of the original data, we evaluate \emph{all} other candidates $(s, \void, a, \void)$ as well, where $a$ is any non-pilot base learner.

\subsubsection{The Filtering Stage (Classifier-Independent Attribute Selection).}
\label{sec:filteringstage}

\begin{figure}[t]
    \centering
    \includegraphics[width=\textwidth]{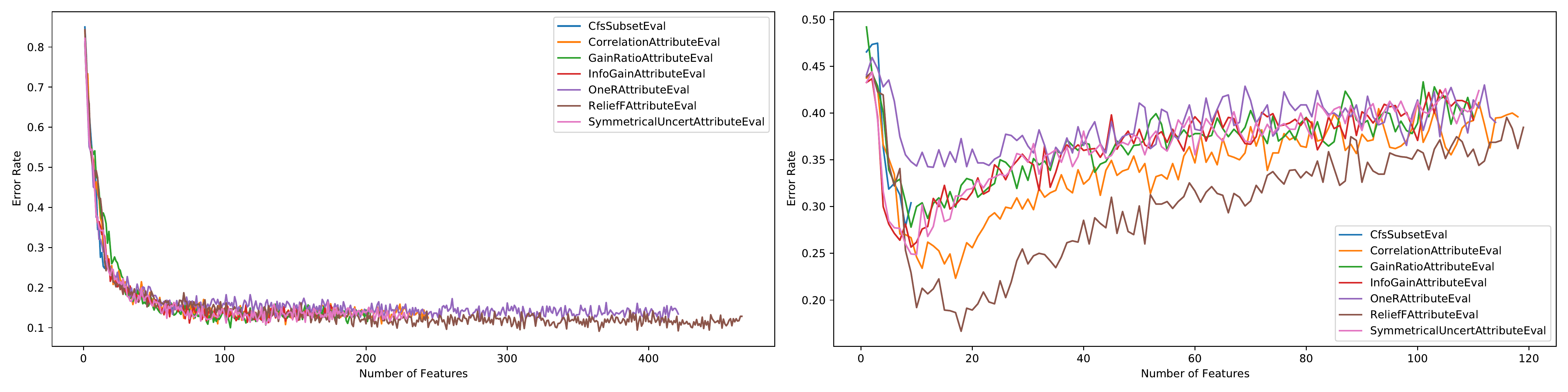}
    \caption{Error rates when applying filtering on the datasets cnae-9 (left) with 856 features in total and Madelon (right) with 500 features in total.
    Each curve shows at point $x$ the error rate obtained when using the first $x$ attributes according to the ranking created by the respective feature evaluator.
    In both cases, it is evident that there is little point in using all the features.
    In the better case, this only implies a waste of computational time (cnae-9) while in the worse case this even implies inferior results (Madelon).
    }
    \label{fig:filtering}
\end{figure}
    This stage consists of two steps.
    In the first step, we compute the set $F$ of features that are considered relevant.
    In the second step, we examine how the previously created candidates behave when using only the features of $F$.
    That is, for every candidate $(s,\void,a,\void)$, we create and evaluate the candidate $(s,F,a,\void)$.
    Here $s$ may be \void as well, and the candidates are created in the order of the performance of the pipeline without feature selection.
    Typically, one may want to define a timeout for this stage to not evaluate overly unpromising candidates if a lot of time has already been used.
    
    The feature set $F$ is computed based on the results of different \emph{filtering} techniques \cite{hall1999correlation}.
    Different techniques to \emph{rank} the relevance of features exist.
    In contrast to \emph{wrapping} \cite{kohavi1997wrappers}, filtering techniques do not adopt (costly) learners to judge feature relevance and hence are typically cheap to compute.
    We execute all such available filters, which gives us a set of rankings.
    Then, for each of these rankings, we compute the performance when using the first $l$ features where $l$ is increased until the performance starts to decrease.
    To assess the performance, a pilot classifier is evaluated on the respective feature set in some kind of cheap cross-validation.
    In this sense, \naml uses existing pre-processors only as a source of \emph{suggestions} for candidate feature sets.

    To motivate this procedure, it is worth to have a look at the \emph{performance curves} one obtains for the different filters.
    For each number $x$ of features, we can plot the performance of some pilot learner if using only the first $x$ features in the ranking created by some filter.
    Fig.~\ref{fig:filtering} illustrates these curves for two datasets and using different filters from WEKA.
    In the case of the cnae-9 dataset (left), we can see that there is only marginal improvement after $x = 100$, and for the madelon dataset (right) we can even observe that performance \emph{decreases} quickly.
    In both cases, it is evident that only a relatively small portion of the features is needed to obtain the same or even better results compared to when all features are used (notice that the plot curves to not even show the full range of features for both datasets).

    Note that the filtering stage optimizes the feature set $F$ independently of the outcomes of earlier stages.
    Neither the results of the probing stage nor those of the scaling stage are used in the determination of $F$.
    Those outcomes are only relevant in that they define the candidates to which the feature selection $F$ should be added, and their order for the evaluation.

    One important additional use case for the analysis of such curves is the reduction of evaluation times during the \automl process.
    Even if we do not \emph{need} to reduce the dimensionality to obtain good results as in the case of madelon (right), we can often still \emph{approximate} the prediction performance on the full feature set sufficiently well. For example, in the case of cnae-9 (left), we only need to use 100 of the 856 features to obtain comparable results, which should be good enough to steer the search process.
    Since we know that the runtime of many learners often increases super-linearly (sometimes quadratically) in the number of attributes \cite{mohr2021runtimeprediction}, there is a huge potential in runtime reduction.
    For example, the time to train a random forest on cnae-9 on some reference machine is 2.2s on average but only 0.5s on 100 attributes with almost the same outcome.

    
    
        
        
    
    
    \subsubsection{Meta-Learner Stage.}
    In this stage, each candidate of the input pool is taken and used as a template to derive new pipelines in which the base learner is wrapped into a meta-learner.
    The feature transformation algorithms, if present, are not touched in this stage.
    
    The algorithms we consider here as meta-learners are also sometimes called (\emph{homogeneous}) ensembles.
    The idea of those learners is to take several copies of a base learner and somehow combine them into a new augmented learner.
    Typical examples are Bagging \cite{breiman1996bagging} and AdaBoost \cite{Freund1998}.
    This is opposed to \emph{heterogeneous} ensembles like Stacking \cite{Wolpert1992} or majority vote ensembles, in which different learners of different types are combined.
    Since we augment existing pipelines (with one learner), we only work with homogeneous ensembles here.

    Forming a heterogeneous ensemble is a reasonable final step.
    This would be identical to the strategy pursued in \cite{feurer2015efficient} to eventually take the best $k$ learners seen so far and merge them into a voting ensemble.
    Alternatively, more sophisticated approaches could try to optimize such an ensemble using previous observations.
    
    
    

    \subsubsection{Parameter Tuning.}
    This stage simply tries to find better hyperparameters for the predictor in one or more candidate pipelines.
    Given a total timeout for this phase, the candidates are optimized in the order of their performance with a (local) timeout and a maximum number of evaluations.
    For algorithms with a small parameter space, all candidates can be enumerated, e.g. k-nearest neighbors with some reasonable candidate set for the number of neighbors.
    For all other algorithms, standard techniques for finding good configurations such as SMAC \cite{hutter2011sequential}, Hyperband \cite{lihyperband2017}, etc. can be employed.
    In this paper, we even only adopt a simple random search.

    \subsubsection{(Validation-Based) Model Selection.}
    \label{sec:validationstage}
    The default decision of \naml to select the best seen pipeline might not always be the best.
    Despite the simplicity and naivity of \naml, there is some significant optimization going on, potentially leading to over-fitting.
    
    The potential need of some validation has been recognized earlier \cite{mohr2018ml}, and we adopt a similar strategy here.
    More precisely, the idea is to keep back a certain portion of the original data that is not shown to the optimization process and \emph{only} used in this final stage.
    The main difference to \cite{mohr2018ml} is that we do not add the validation data to the pool and then run a cross-validation on the augmented dataset, but we here simply conduct a ``classical'' single-fold validation on this hold-out set.
    
    The data portion used for validation can be chosen dynamically based on the desired guarantees on the generalization performance.
    For a set of $m$ remaining competitive candidates, the Hoeffding bound allows us to estimate the out-of-sample error with $\mathbb{P}(|\mu - \nu| > \varepsilon) < 2me^{-2\varepsilon n}$, where $\mu$ is the true out-of-sample error, $\nu$ is the error on the validation fold, and $n$ is the \emph{size} of the validation fold.
    Given sufficient data and for a moderate number $m$, say $m = 10$, this can be quite a good bound and impose an important remedy against over-fitting after an exhaustive optimization effort.
    
    Unfortunately, the number of validation samples available is often not sufficient to make the Hoeffding bound meaningful.
    In most cases, we want to choose $\varepsilon \approx 0.01$ and have the bound relatively small, say, $0.1$.
    To assure such a bound for even only one candidate (effectively \emph{testing} its performance), the validation fold must already contain roughly 15000 examples.
    For $m = 10$, we would need 35,000 validation instances, which are often not available.
    
    We hence propose to use both the ``internal'' score observed during the optimization process \emph{and} the validation score for model selection and weight the two scores based on the validation set size.
    To this end, we introduce a parameter $\bar{n}$ that quantifies the number of validation instances required to exclusively use validation instances to compute the score.
    Intuitively, $\bar{n}$ is the number of instances needed to get the desired certainty in the Hoeffding bound, e.g., 35,000 instances.
    The \emph{satisfaction} of the Hoeffding bound can then be expressed as $$\tau(n) = \min\{1, \frac{n}{\bar{n}}\},$$
    where $n$ is the actual size of the validation fold.
    However, a close-to-0 satisfaction does not necessarily mean a close-to-0 weight of the validation score.
    For example, if we have 500 instances in total and use 100 of them for validation, then 100 is probably far away from $\bar{n}$, and the satisfaction is almost 0.
    Still, the 100 instances are a valuable complement to the 400 instances used for training.
    In this case, it makes more sense to weight the validation score based on the \emph{ratio} between the number of validation instances and totally available instances; here, this would be 0.2.
    Hence, instead of using the satisfaction of the validation fold size directly as a weight, we use it to determine the point on a linear scale between the above ratio and 1.
    Formally, we define the pipeline score in the validation phase then to be
    $$\performance_{final}(P) = \performance_{int}(P)(1 - \omega(n)) + \performance_{validate}(P)\omega(n),$$
    where $\performance_{int}$ is the internal score obtained using only the data available for optimization, $\performance_{validate}$ is the performance obtained using \emph{only} the validation data (the pipeline has still been defined not using these data points), $n$ is the number of instances in the validation fold, and
    $$\omega(n) = \tau(n) + \frac{n}{N}(1-\tau(n)).$$
    In the last term, $N$ is the total number of instances available.
    

\section{Evaluation}
\label{sec:evaluation}
In this section, we want to address the following research questions:
\begin{enumerate}
    
    \item How much can state-of-the-art tools for Python and Java improve over the baseline imposed by \naml?
    
    \item What is the independent merit of each technique over the probing stage?
    \item Are there any synergetic effects observable when employing several stages?
\end{enumerate}

We stress our point of view that \naml is in fact the baseline here and not the competing technique.
Once more we argue that the baselines proposed in \cite{autoweka} such as \exdef or a grid search do not reflect an expert data scientist.
\naml hence \emph{updates} this baseline by proposing a more sophisticated data scientist (acknowledging that an expert data scientist would be even more flexible and hence might even be stronger than our \naml).
In this sense, our evaluation could be understood as a way of showing how experimental evaluations of other tools might have looked like if a more sophisticated baseline, like \naml, would have been adopted.

\subsection{Compared Algorithms}
To answer the above questions, we ran \naml in several configurations.
First, we computed results (and runtimes) of \naml when using no or at most one of the optional stages.
This allows us to see whether a particular stage alone can improve over the primitive selection algorithm.
In addition, we considered the \emph{monotone} scenarios in which we consider stage setups of increasing ``complexity'' (scenario $i$ takes all stages prior to $i$ and \emph{adds} stage $i$ to this set).
This allows us to observe whether the \emph{combination} of two techniques can improve upon the simpler approach.

In the following, we shortcut the two extreme profiles to give a summarized comparison.
We refer to ``primitive'' as the version of \naml that \emph{only} adopts the probing stage and nothing else.
Once again, this corresponds to the \exdef baseline proposed in \cite{autoweka}.
The profile containing \emph{all} the stages (including validation), is referred to as ``full''.
The validation stage, if applied sets $\bar{n} = 10000$ and considers the best 10 candidates in the pool.

On the state-of-the-art side, we compare solutions with competitive performance for both WEKA \cite{hall2009weka} and scikit-learn \cite{pedregosa2011scikit}.
It is well-known that \autoweka is not competitive even inside the WEKA domain \cite{mohr2018ml}, but we still consider it here to contrast our results of the comparison of \automl with the \exdef baseline to those reported in \cite{autoweka}.
To our knowledge, \mlplan \cite{mohr2018ml} is the best performing \automl tool for WEKA to date, so we consider it as the second baseline for WEKA.
On the scikit-learn side, we consider only \autosklearn as a baseline (without warm-starting and without final ensemble building).
While \autosklearn has been beaten by several other tools, these tools either rely on warm-starting (\cite{yang2019oboe,fusi2018probabilistic}) or do not outpferform \autosklearn to such a degree that we would have to consider it a second-class solution.
The latter can be either due to the fact that results are generally comparable, like in the case of TPOT \cite{olson2016tpot}, or that results are only reported in the way of average ranks, which obscure the factual improvement \cite{DBLP:conf/ijcai/RakotoarisonSS19,liu2020admm}.
Hence, \autosklearn is still a state-of-the-art solution to us.
In this paper, we consider version 0.12.0, which underwent substantial changes and improvements compared to the original version \cite{feurer2015efficient}.

\subsection{Experiment Setup}
The evaluation is based on the dataset portfolio proposed in \cite{gijsbers2019open}.
This is a collection of datasets available on openml.org \cite{OpenML2013}.
To complement these datasets, we have added some of the datasets that were used in the \autoweka paper \cite{autoweka} and have been used frequently for comparison in publications on \automl.
Five datasets of \cite{gijsbers2019open} (23, 31, 188, 40996, 41161, 42734) were removed due to technical issues in the data loading process.

For each dataset, 10 random train-test splits were created, and all algorithms were run once on each of these splits, using the train data for optimizing, and the test data to assessing the performance.
Needless to say, the splits were the same for all approaches.

Timeouts were configured as follows.
For all algorithms, we allowed a total runtime of 1h, and the runtime for a single pipeline execution was configured to take up to 1 minute.
In \autoweka, it is not possible to configure the maximum runtime for single executions; this variable is interenally controlled.
In \naml, we imposed stage time bounds for the meta and the parameter tuning stages of 5 minutes respectively (the other stages were not equipped with a dedicated timeout).
Of course, on hitting the time bound of 1h, \naml was stopped regardless the phase in which it was, and the best seen solution was returned.

The computations were executed in a compute center with Linux machines, each of them equipped with 2.6Ghz Intel Xeon E5-2670 processors and 32GB memory.
In spite of the technical possibilities, we did \emph{not} parallelize evaluations.
That is, all the tools were configured to run with a single CPU core\footnote{This was mainly due to technical reasons, because the standard parallelization mechanism in Python does not support shared memory and copies contexts by default to the different threads. This can blow-up the memory for some of the considered datasets, so we refrained from parallelism.}.
Still, \naml has immediate support for parallelization, and the code (both Java and Python) that was used for the experiments is publicly available\footnote{ \url{https://github.com/fmohr/naiveautoml}}.

\subsection{Results}

\begin{table}[t]
    \centering
    \caption{Mean error rates and standard deviations per machine learning library.}
    \label{tab:mainresults}
    \resizebox{\textwidth}{!}{
\begin{tabular}{r||rr|rr||r|rr}
\toprule
 & \multicolumn{4}{c||}{WEKA backend} & \multicolumn{3}{c}{scikit-learn backend}\\
 \multicolumn{1}{c}{id} &                                  \multicolumn{1}{c}{auto-weka} &                                     \multicolumn{1}{c}{mlplan} & \multicolumn{1}{c}{primitive} &   \multicolumn{1}{c}{full} &                                   \multicolumn{1}{c}{asklearn} & \multicolumn{1}{c}{primitive} &   \multicolumn{1}{c}{full} \\
\midrule
                      3 &       \textbf{0.0$\pm$0.0} \phantom{$\circ$} \phantom{$\circ$} &  \underline{0.01$\pm$0.01} \phantom{$\circ$} \phantom{$\circ$} &          \textbf{0.0$\pm$0.0} &       \textbf{0.0$\pm$0.0} &       \textbf{0.0$\pm$0.0} \phantom{$\circ$} \phantom{$\circ$} &          \textbf{0.0$\pm$0.0} &       \textbf{0.0$\pm$0.0} \\
                     12 &      \textbf{0.02$\pm$0.0} \phantom{$\circ$} \phantom{$\circ$} &              0.04$\pm$0.01 \phantom{$\circ$} \phantom{$\circ$} &     \underline{0.03$\pm$0.01} &  \underline{0.03$\pm$0.01} &  \underline{0.03$\pm$0.02} \phantom{$\circ$} \phantom{$\circ$} &        \textbf{0.02$\pm$0.02} &  \underline{0.03$\pm$0.02} \\
                     54 &  \underline{0.18$\pm$0.02} \phantom{$\circ$} \phantom{$\circ$} &  \underline{0.18$\pm$0.03} \phantom{$\circ$} \phantom{$\circ$} &        \textbf{0.17$\pm$0.02} &  \underline{0.18$\pm$0.05} &                                  0.19$\pm$0.03 $\circ$ $\circ$ &        \textbf{0.16$\pm$0.03} &  \underline{0.17$\pm$0.03} \\
                    181 &  \underline{0.39$\pm$0.04} \phantom{$\circ$} \phantom{$\circ$} &  \underline{0.39$\pm$0.04} \phantom{$\circ$} \phantom{$\circ$} &        \textbf{0.38$\pm$0.03} &  \underline{0.39$\pm$0.04} &  \underline{0.39$\pm$0.04} \phantom{$\circ$} \phantom{$\circ$} &     \underline{0.38$\pm$0.03} &     \textbf{0.37$\pm$0.05} \\
\bottomrule
\end{tabular}

    }
\end{table}
The high level results of the experiments are shown in Table \ref{tab:mainresults}.
The reported metric is the \emph{error rate}.
For each dataset and each approach, we report the \emph{trimmed} mean (10\% trimmed) with standard deviation.
Nan entries are caused by memory overflows.
We refrain from the now somewhat common practice of reporting average ranks, because, in our view, those obscure a lot of important details in the comparison, e.g. on \emph{which} (and \emph{how many}) datasets which algorithm is better than another and by which \emph{margin}.

The formatting semantics of the table is as follows.
The best entries (w.r.t to the trimmed mean) per machine learning library are formatted in bold, and those whose result distribution cannot be said to be statistically different (according to a Wilcoxcon signed rank test with confidence 0.05) or where the performance difference is ``irrelevant'' ($< 0.01$) are underlined.
With respect to the latter one we of course understand that in some cases such marginal difference \emph{can} be relevant, but here we still treat them as equally good to give more significance to the symbols used to denote improvements.
For each of the state-of-the-art techniques, we use the $\bullet$ or $\circ$ symbol to indicate a score that is ``substantially'' better or worse than the one obtained with \naml.
The symbols are used twice, once to compare against the ``primitive'' profile and once comparing against the ``full'' profile.
As above, substantial here means that it is not only statistically significant but also that the absolute improvement is at least 0.01.
The symbol $\dagger$ is used instead of a $\bullet$ in cases in which \autosklearn constructed pipelines with feature transformation algorithms not supported in the algorithm scheme of \naml; examples are the PCA or feature encodings based on trees.
Based on these results, we can now answer the three research questions above.

\subsubsection{RQ 1: Performance of State-Of-The-Art Approaches over \naml}
When asking how much the currently best \automl approaches can improve over the \naml baseline, the answer contained in Table \ref{tab:mainresults} is, put mildly, surprising.
While one would expect \naml to be frequently outperformed by the other approaches, we cannot observe any such type of dominance.
In fact, there are even quite some datasets on which \naml performs \emph{better}.

What is more, the above results seem to contradict the first (and only) comparison of an \automl tool against a simulated data scientist presented in \cite{autoweka}.
In that paper, \autoweka systematically outperforms the \exdef baseline, which here corresponds to the performance in the ``primitive'' column.
On \emph{no} dataset is \autoweka better than this baseline and is even sometimes outperformed by it.

This quite striking observation can also be put into words by saying that if we run a simple \texttt{for} loop over the possible base learners (using default parameters and ignoring meta-learners or any kind of feature transformation), then we obtain an equal or better performance than \autoweka, \mlplan, or \autosklearn in more than 90\% of the cases.
This somewhat shocking observation alone strongly jeopardizes the justification to apply resource-intense \automl tools.
As shown in the appendix, the results for the ``primitive'' solution arrive often in the range of some few seconds or at least some minutes.
So they are cheap to obtain, and other approaches rarely can improve upon it even when run for an hour or more.
This means that the least we should expect from evaluations of new \automl approaches is that they come back to compare against the \exdef baseline.

When considering the ``full'' baseline, the situation looks similar.
Clearly, since ``primitive'' is already such a strong baseline, we would not expect the ``full'' baseline to improve upon that too often.
However, in some cases, the ``full'' baseline performs better than the ``primitive'' profile.
This can almost always be attributed to advantages gained in the feature scaling or feature selection.
Maybe surprisingly, sometimes the opposite is the case and the ``full'' profile performs worse than the ``primitive'' one.
We discuss this in more detail in Sec. \ref{sec:results:merits} and \ref{sec:results:synergies}, but the short explanation is that ``full'' can be weaker than ``primitive'' because it includes the \emph{validation} stage.
The latter implies that less data is used to find a model, which
is a drawback in some situations.

In our view, there are two possible explanations for the missing superiority of the \blackbox optimization techniques.
The first is simply that the global optimum \emph{is} in fact obtained by choosing the best algorithm without any further optimizations.
That is, there is simply no \emph{potential} for optimization.
On some datasets (with almost perfect performance) this is clearly the case.
The second is that there \emph{maybe} is potential for optimization, but the resource limitations impede that the \blackbox approaches can develop their full potential.
Theoretically, all of the \blackbox approaches converge to the globally optimal solution.
However, only a tiny fraction of the search space can be examined in any reasonable timeout, and the number of evaluations that can be made in that time are not enough to learn enough about the performance landscape to steer the search process in a meaningful way.
In other words, the \blackbox optimizers \emph{might} be better in theory and in the limit, but the limitations imposed by the time constraints impede that any of the algorithms substantially leaves the exploration phase and can actually exploit the model it learnt.

\subsubsection{RQ 2: Individual Merits of the Optimization Stages.}
\label{sec:results:merits}
To answer the question which of the stages is responsible for improvements over the ``primitive'' profile, we create a tournament table.
The appendix contains the full result tables for the different stages.
For each stage, we check how often a substantial improvement over the primitive stage is obtained (win), how often this degree of improvement can be achieved \emph{only} with this stage for the respective dataset (unique win), how often a substantial degradation can be observed (loss), and how often there is just no change (draw).

\begin{center}
    \resizebox{\textwidth}{!}{
        \begin{tabular}{l|rrrr||rrrrr}
\toprule
 & \multicolumn{4}{c||}{WEKA backend} & \multicolumn{5}{c}{scikit-learn backend}\\
{} &  filtering &  meta &  tuning &  validation &  scaling &  filtering &  meta &  tuning &  validation \\
\midrule
wins        &          6 &     4 &       1 &           4 &        7 &          2 &     2 &       3 &           1 \\
unique wins &          5 &     1 &       0 &           3 &        7 &          2 &     0 &       0 &           1 \\
losses      &          2 &     5 &       4 &           4 &        1 &          1 &     1 &       2 &           1 \\
draws       &         53 &    52 &      56 &          53 &       53 &         58 &    58 &      56 &          59 \\
\bottomrule
\end{tabular}

    }
\end{center}

This summary provides several interesting insights.
The first observation is that, in the majority of cases, none of the stages has any effect at all onto the overall result.
Even in the case of \emph{scaling} in scikit-learn, which presents the highest number of changes compared to the \emph{primitive} setup, less than 15\% of the cases exhibit a substantial change in the results.
However, we can also observe that at least there  is the potential to improve upon the probing stage in some cases.

What is also interesting is that many of the improvements are \emph{unique} to one of the techniques.
For example, the scaling stage in scikit-learn achieved wins over the probing stage on 7 datasets, and on all of those datasets, \emph{only} the scaling stage was able to improve upon the probing stage.
Summarizing then the row of unique wins, we can assert that out of the 61 examined datasets, we could improve upon 9 datasets with \emph{exactly} one of the stages in WEKA and on 10 in the case of scikit-learn.
This result highlights the necessity of complementary stages, and observing the fact that we left out a stage for feature transformation that was responsible for even better performance in \autosklearn (the $\dagger$ cases), this effect can be expected to be even bigger when more types of algorithms are considered.

\subsubsection{RQ 3: Synergies Between Stages}
\label{sec:results:synergies}
We analyze this question in a similar fashion as the previous one via a tournament table.
We now not consider only a single stage but a \emph{range} of stages (starting with the probing stage and including \emph{all} stages up to the one of interest).
For example, the stage range defined by \emph{tuning} include (i) probing, (ii) scaling, (iii) filtering, (iv) meta, and (v) tuning.
From this logic, there are four relevant stage ranges; the scaling stage range is the same as reported above and is hence omitted.
For each such range of stages, we now count the number of datasets on which there was an improvement over the probing stage that was \emph{better} than the improvement achieved by \emph{any} of the stage contained in the range alone (reported above).

\begin{center}
    \resizebox{.8\textwidth}{!}{
    \begin{tabular}{l||rrr|rrrr}
\toprule
 & \multicolumn{3}{c|}{WEKA backend} & \multicolumn{3}{c}{scikit-learn backend}\\
{} &  meta &  tuning &  validation &  filtering &  meta &  tuning &  validation \\
\midrule
wins   &     0 &       0 &           0 &          0 &     2 &       0 &           0 \\
losses &     0 &       0 &           0 &          2 &     4 &       7 &           8 \\
draws  &    61 &      61 &          61 &         59 &    55 &      54 &          53 \\
\bottomrule
\end{tabular}
    }
\end{center}

The results show that there are \emph{no} (positive) synergies between the stages.
While this may seem a bit natural for the filtering stage range (because up to there, there is essentially no communication between the stages), it may be surprising for the other stages.
The absence of synergies between stages in \emph{this} evaluation can be seen as an \emph{indicator} that the naivity assumption is indeed justified.
However and once more, we do not even claim that the naivity assumptions are justified in general but only aim at giving evidence that in \emph{many} situations, the results obtained with a naive approach are competitive.
In our view, the performance of the state-of-the-art tools is not particularly challenging this simple view.

Note that the above results do of course \emph{not} mean that we can run \naml with just one of the stages.
What we learn is that if there is an improvement over the probing stage, then this improvement can be attributed to one of the stages.
However, we do not know in advance which of the stages that could be.
Since most stages imply the same number of deteriorations as of improvements, it seems indeed best to only apply the scaling and filtering stage and omit meta-learners, tuning and validation.

In this whole evaluation, the effect of the validation stage is quite disappointing.
It seems natural to adscribe the deteriorations in the earlier stages to an over-fitting effect, and one would hope to combat this over-fitting in the validation stage.
Apparently, this does not work in the desired way.
A nearby explanation is that the reduction in data available for the optimization process misleads the optimization and proposes sub-optimal models, and this effect cannot be compensated by validation later.
However, there might be also other reasons, and studying this effect in more detail may be interesting future work.

\section{Conclusion}
This paper proposes \naml, a slightly more sophisticated baseline for \automl than proposed with \exdef in \cite{autoweka}.
Instead of searching a complex pipeline search space with a \blackbox optimizer, \naml imitates the sequential workflow conducted by a data scientist, which \emph{implicitly} defines a solution pipeline.
We empirically demonstrate that state-of-the-art tools are not able to substantially outperform this baseline (sometimes the contrary is true).
Given the increasing demand of transparent and understandable \automl \cite{wang2019human,wang2021autods,drozdal2020trust}, this result highly jeopardizes the approach of using \blackbox optimization and favors the idea of imitating the data scientist's workflow.
Producing competitive results, this process is not only much more transparent and understandable to the expert---and hence allows better interaction between the expert and the machine---, but also is the approach much more flexible, because all modifications can be directly realized in the stage implementations instead of having it to be injected into the solver through the problem encoding.

Our suggestions for future work go hence also into the direction of extending and improving the naive approach instead of spending more time on enhancements of \blackbox optimization approaches.
Besides the obvious option to consider other stage schemes and more complex pipelines (with tree-shaped pre-processing), interesting future work lies in the detection of stage saturation: When can we be sure that a stage will not improve upon the currently best known solution?
Also, creating a more interactive version of \naml in which the expert obtains visual summaries of what has been done and with the option to stop a particular stage and pass to the next seem highly relevant.

\printbibliography

@article{lihyperband2017,
  author    = {Lisha Li and
               Kevin G. Jamieson and
               Giulia DeSalvo and
               Afshin Rostamizadeh and
               Ameet Talwalkar},
  title     = {Hyperband: {A} Novel Bandit-Based Approach to Hyperparameter Optimization},
  journal   = {J. Mach. Learn. Res.},
  volume    = {18},
  pages     = {185:1--185:52},
  year      = {2017},
  url       = {http://jmlr.org/papers/v18/16-558.html},
  bibsource = {dblp computer science bibliography, https://dblp.org}
}

@inproceedings{chen2018autostacker,
  title={Autostacker: A compositional evolutionary learning system},
  author={Chen, Boyuan and Wu, Harvey and Mo, Warren and Chattopadhyay, Ishanu and Lipson, Hod},
  booktitle={Proceedings of the Genetic and Evolutionary Computation Conference},
  pages={402--409},
  year={2018}
}

@inproceedings{yang2019oboe,
  title={OBOE: Collaborative filtering for AutoML model selection},
  author={Yang, Chengrun and Akimoto, Yuji and Kim, Dae Won and Udell, Madeleine},
  booktitle={Proceedings of the 25th ACM SIGKDD International Conference on Knowledge Discovery \& Data Mining},
  pages={1173--1183},
  year={2019}
}

@book{boyd2011distributed,
  title={Distributed optimization and statistical learning via the alternating direction method of multipliers},
  author={Boyd, Stephen and Parikh, Neal and Chu, Eric},
  year={2011},
  publisher={Now Publishers Inc}
}

@inproceedings{feurer2015efficient,
  title={Efficient and robust automated machine learning},
  author={Feurer, Matthias and Klein, Aaron and Eggensperger, Katharina and Springenberg, Jost and Blum, Manuel and Hutter, Frank},
  booktitle={Advances in Neural Information Processing Systems},
  pages={2962--2970},
  year={2015}
}

@inproceedings{autoweka,
  author    = {Chris Thornton and Frank Hutter and Holger H.\ Hoos and Kevin Leyton-Brown},
  title     = {Auto-{WEKA}: combined selection and hyperparameter optimization of classification algorithms},
  booktitle = {The 19th {ACM} {SIGKDD} International Conference on Knowledge Discovery and Data Mining, {KDD} 2013, Chicago, IL, USA},
  pages     = {847--855},
  year      = {2013}
}

@inproceedings{liu2020admm,
  title={An ADMM based framework for automl pipeline configuration},
  author={Liu, Sijia and Ram, Parikshit and Vijaykeerthy, Deepak and Bouneffouf, Djallel and Bramble, Gregory and Samulowitz, Horst and Wang, Dakuo and Conn, Andrew and Gray, Alexander},
  booktitle={Proceedings of the AAAI Conference on Artificial Intelligence},
  volume={34},
  number={04},
  pages={4892--4899},
  year={2020}
}

@inproceedings{fusi2018probabilistic,
  title={Probabilistic matrix factorization for automated machine learning},
  author={Fusi, Nicolo and Sheth, Rishit and Elibol, Melih},
  booktitle={Proceedings of the 32nd International Conference on Neural Information Processing Systems},
  pages={3352--3361},
  year={2018}
}

@article{hutter2011sequential,
  title={Sequential Model-Based Optimization for General Algorithm Configuration.},
  author={Hutter, Frank and Hoos, Holger H and Leyton-Brown, Kevin},
  journal={LION},
  volume={5},
  pages={507--523},
  year={2011},
  publisher={Springer}
}

@inproceedings{olson2016tpot,
  title={TPOT: A Tree-based Pipeline Optimization Tool for Automating Machine Learning},
  author={Olson, Randal S and Moore, Jason H},
  booktitle={Workshop on Automatic Machine Learning},
  pages={66--74},
  year={2016}
}

@article{nguyen2014using,
  title={Using meta-mining to support data mining workflow planning and optimization},
  author={Nguyen, Phong and Hilario, Melanie and Kalousis, Alexandros},
  journal={Journal of Artificial Intelligence Research},
  volume={51},
  pages={605--644},
  year={2014}
}

@inproceedings{kietz2012designing,
  title={Designing {KDD}-Workflows via {HTN}-Planning for Intelligent Discovery Assistance},
  author={Kietz, J{\"o}rg-Uwe and Serban, Floarea and Bernstein, Abraham and Fischer, Simon},
  booktitle={5th Planning to Learn Workshop WS28 at ECAI 2012},
  pages={10},
  year={2012}
}

@inproceedings{de2017recipe,
  title={RECIPE: a grammar-based framework for automatically evolving classification pipelines},
  author={de S{\'a}, Alex GC and Pinto, Walter Jos{\'e} GS and Oliveira, Luiz Otavio VB and Pappa, Gisele L},
  booktitle={European Conference on Genetic Programming},
  pages={246--261},
  year={2017},
  organization={Springer}
}

@article{OpenML2013,
author = {Vanschoren, Joaquin and van Rijn, Jan N. and Bischl, Bernd and Torgo, Luis},
title = {{OpenML}: Networked Science in Machine Learning},
journal = {SIGKDD Explorations},
volume = {15},
number = {2},
year = {2013},
pages = {49--60},
url = {http://doi.acm.org/10.1145/2641190.2641198},
doi = {10.1145/2641190.2641198},
publisher = {ACM},
address = {New York, NY, USA},
}

@article{kotthoff2017auto,
  title={Auto-WEKA 2.0: Automatic model selection and hyperparameter optimization in WEKA},
  author={Kotthoff, Lars and Thornton, Chris and Hoos, Holger H and Hutter, Frank and Leyton-Brown, Kevin},
  journal={The Journal of Machine Learning Research},
  volume={18},
  number={1},
  pages={826--830},
  year={2017},
  publisher={JMLR. org}
}

@inproceedings{kietz2009towards,
  title={Towards cooperative planning of data mining workflows},
  author={Kietz, J and Serban, Floarea and Bernstein, Abraham and Fischer, Simon},
  booktitle={Proceedings of the Third Generation Data Mining Workshop at the 2009 European Conference on Machine Learning},
  pages={1--12},
  year={2009},
  organization={Citeseer}
}

@inproceedings{nguyen2011meta,
  title={A meta-mining infrastructure to support kd workflow optimization},
  author={Nguyen, Phong and Kalousis, Alexandros and Hilario, Melanie},
  booktitle={Proceedings of  the PlanSoKD-11 Workshop at ECML/PKDD},
  pages={1--10},
  year={2011}
}

@inproceedings{nguyen2012experimental,
  title={Experimental evaluation of the e-lico meta-miner},
  author={Nguyen, Phong and Kalousis, Alexandros and Hilario, Melanie},
  booktitle={5th planning to learn workshop WS28 at ECAI},
  pages={18--19},
  year={2012}
}

@inproceedings{DBLP:conf/ijcai/RakotoarisonSS19,
  author    = {Herilalaina Rakotoarison and
               Marc Schoenauer and
               Mich{\`{e}}le Sebag},
  editor    = {Sarit Kraus},
  title     = {Automated Machine Learning with Monte-Carlo Tree Search},
  booktitle = {Proceedings of the Twenty-Eighth International Joint Conference on
               Artificial Intelligence, {IJCAI} 2019, Macao, China, August 10-16,
               2019},
  pages     = {3296--3303},
  publisher = {ijcai.org},
  year      = {2019},
  url       = {https://doi.org/10.24963/ijcai.2019/457},
  doi       = {10.24963/ijcai.2019/457},
  bibsource = {dblp computer science bibliography, https://dblp.org}
}

@article{hall2009weka,
  title={The WEKA data mining software: an update},
  author={Hall, Mark and Frank, Eibe and Holmes, Geoffrey and Pfahringer, Bernhard and Reutemann, Peter and Witten, Ian H},
  journal={ACM SIGKDD Explorations},
  volume={11},
  _number={1},
  _pages={10--18},
  year={2009},
  _publisher={ACM New York, NY, USA}
}

@article{pedregosa2011scikit,
  title={Scikit-learn: Machine learning in Python},
  author={Pedregosa, Fabian and Varoquaux, Ga{\"e}l and Gramfort, Alexandre and Michel, Vincent and Thirion, Bertrand and Grisel, Olivier and Blondel, Mathieu and Prettenhofer, Peter and Weiss, Ron and Dubourg, Vincent and others},
  journal={the Journal of machine Learning research},
  volume={12},
  pages={2825--2830},
  year={2011},
  publisher={JMLR. org}
}

@article{gijsbers2019open,
  title={An open source automl benchmark},
  author={Gijsbers, Pieter and LeDell, Erin and Thomas, Janek and Poirier, S{\'e}bastien and Bischl, Bernd and Vanschoren, Joaquin},
  journal={arXiv preprint arXiv:1907.00909},
  year={2019}
}

@article{mohr2018ml,
  title={ML-Plan: Automated machine learning via hierarchical planning},
  author={Mohr, Felix and Wever, Marcel and H{\"u}llermeier, Eyke},
  journal={Machine Learning},
  volume={107},
  number={8},
  pages={1495--1515},
  year={2018},
  publisher={Springer}
}

@article{breiman1996bagging,
  title={Bagging predictors},
  author={Breiman, Leo},
  journal={Machine learning},
  volume={24},
  _number={2},
  _pages={123--140},
  year={1996},
  publisher={Springer}
}

@inproceedings{Freund1998,
    address = {New York, NY},
    author = {Y. Freund and R. E. Schapire},
    booktitle = {11th Annual Conference on Computational Learning Theory},
    pages = {209-217},
    publisher = {ACM Press},
    title = {Large margin classification using the perceptron algorithm},
    year = {1998}
 }

@article{Wolpert1992,
    author = {David H. Wolpert},
    journal = {Neural Networks},
    pages = {241-259},
    publisher = {Pergamon Press},
    title = {Stacked generalization},
    volume = {5},
    year = {1992}
}

@article{crisan2021fits,
  title={Fits and Starts: Enterprise Use of AutoML and the Role of Humans in the Loop},
  author={Crisan, Anamaria and Fiore-Gartland, Brittany},
  journal={arXiv preprint arXiv:2101.04296},
  year={2021}
}

@article{wang2021autods,
  title={AutoDS: Towards Human-Centered Automation of Data Science},
  author={Wang, Dakuo and Andres, Josh and Weisz, Justin and Oduor, Erick and Dugan, Casey},
  journal={arXiv preprint arXiv:2101.05273},
  year={2021}
}

@inproceedings{drozdal2020trust,
  title={Trust in automl: Exploring information needs for establishing trust in automated machine learning systems},
  author={Drozdal, Jaimie and Weisz, Justin and Wang, Dakuo and Dass, Gaurav and Yao, Bingsheng and Zhao, Changruo and Muller, Michael and Ju, Lin and Su, Hui},
  booktitle={Proceedings of the 25th International Conference on Intelligent User Interfaces},
  pages={297--307},
  year={2020}
}

@article{wang2019human,
  title={Human-AI collaboration in data science: Exploring data scientists' perceptions of automated AI},
  author={Wang, Dakuo and Weisz, Justin D and Muller, Michael and Ram, Parikshit and Geyer, Werner and Dugan, Casey and Tausczik, Yla and Samulowitz, Horst and Gray, Alexander},
  journal={Proceedings of the ACM on Human-Computer Interaction},
  volume={3},
  number={CSCW},
  pages={1--24},
  year={2019},
  publisher={ACM New York, NY, USA}
}

@thesis{hall1999correlation,
  title={Correlation-based feature selection for machine learning},
  author={Hall, Mark Andrew},
  year={1999},
  publisher={University of Waikato Hamilton}
}

@article{kohavi1997wrappers,
  title={Wrappers for feature subset selection},
  author={Kohavi, Ron and John, George H},
  journal={Artificial intelligence},
  volume={97},
  number={1-2},
  pages={273--324},
  year={1997},
  publisher={Elsevier}
}

@article{mohr2021runtimeprediction,
  author={F. {Mohr} and M. {Wever} and A. {Tornede} and E. {H\"ullermeier}},
  journal={IEEE Transactions on Pattern Analysis and Machine Intelligence}, 
  title={Predicting Machine Learning Pipeline Runtimes in the Context of Automated Machine Learning}, 
  year={2021},
  volume={},
  number={},
  pages={1-1},
  doi={10.1109/TPAMI.2021.3056950}
}

\newpage
\appendix
\section*{Appendix}
\section{Datasets}
All datasets are available via the openml.org platform \cite{OpenML2013}.
\begin{table}[h!]
    \centering
    \resizebox{.7\textwidth}{!}{
\begin{tabular}{rlrrr}
\toprule
 openmlid &                                    name &  instances &  features &  classes \\
\midrule
        3 &                                kr-vs-kp &       3196 &        37 &        2 \\
       12 &                           mfeat-factors &       2000 &       217 &       10 \\
       54 &                                 vehicle &        846 &        19 &        4 \\
      181 &                                   yeast &       1484 &         9 &       10 \\
     1049 &                                     pc4 &       1458 &        38 &        2 \\
     1067 &                                     kc1 &       2109 &        22 &        2 \\
     1111 &                      KDDCup09\_appetency &      50000 &       231 &        2 \\
\bottomrule
\end{tabular}

    }
    \caption{Overview of datasets used in the evaluation.}
    \label{tab:datasets}
\end{table}

\section{Result Tables for \naml}
The following tables contain the results for the different stages and the respective comparison to the state-of-the-art approaches.

\begin{table}[h!]
    \centering
    \resizebox{\textwidth}{!}{
        \begin{tabular}{rrrrrrrr}
\toprule
 openmlid &                                                                                                            auto-weka &                                                                                                               mlplan &                  primitive &                  filtering &                       meta &                     tuning &                 validation \\
\midrule
        3 &       \textbf{0.0$\pm$0.0} \phantom{$\circ$} \phantom{$\circ$} \phantom{$\circ$} \phantom{$\circ$} \phantom{$\circ$} &  \underline{0.01$\pm$0.01} \phantom{$\circ$} \phantom{$\circ$} \phantom{$\circ$} \phantom{$\circ$} \phantom{$\circ$} &       \textbf{0.0$\pm$0.0} &       \textbf{0.0$\pm$0.0} &       \textbf{0.0$\pm$0.0} &       \textbf{0.0$\pm$0.0} &       \textbf{0.0$\pm$0.0} \\
       12 &      \textbf{0.02$\pm$0.0} \phantom{$\circ$} \phantom{$\circ$} \phantom{$\circ$} \phantom{$\circ$} \phantom{$\circ$} &              0.04$\pm$0.01 \phantom{$\circ$} \phantom{$\circ$} \phantom{$\circ$} \phantom{$\circ$} \phantom{$\circ$} &  \underline{0.03$\pm$0.01} &  \underline{0.03$\pm$0.01} &  \underline{0.03$\pm$0.01} &  \underline{0.03$\pm$0.01} &  \underline{0.03$\pm$0.01} \\
       54 &  \underline{0.18$\pm$0.02} \phantom{$\circ$} \phantom{$\circ$} \phantom{$\circ$} \phantom{$\circ$} \phantom{$\circ$} &  \underline{0.18$\pm$0.03} \phantom{$\circ$} \phantom{$\circ$} \phantom{$\circ$} \phantom{$\circ$} \phantom{$\circ$} &     \textbf{0.17$\pm$0.02} &     \textbf{0.17$\pm$0.02} &  \underline{0.18$\pm$0.02} &     \textbf{0.17$\pm$0.02} &  \underline{0.18$\pm$0.04} \\
\bottomrule
\end{tabular}

    }
    \caption{Stage-Wise Results in Isolated Configuration for WEKA.}
    \label{tab:namlresults:monotone:python}
\end{table}

\newpage
\begin{table}[h!]
    \centering
    \resizebox{\textwidth}{!}{
        \begin{tabular}{rrrrrrrr}
\toprule
 openmlid &                                                                                                                               asklearn &                  primitive &                    scaling &                  filtering &                       meta &                     tuning &                 validation \\
\midrule
        3 &       \textbf{0.0$\pm$0.0} \phantom{$\circ$} \phantom{$\circ$} \phantom{$\circ$} \phantom{$\circ$} \phantom{$\circ$} \phantom{$\circ$} &       \textbf{0.0$\pm$0.0} &       \textbf{0.0$\pm$0.0} &       \textbf{0.0$\pm$0.0} &                        nan &       \textbf{0.0$\pm$0.0} &      \textbf{0.0$\pm$0.01} \\
       12 &  \underline{0.03$\pm$0.02} \phantom{$\circ$} \phantom{$\circ$} \phantom{$\circ$} \phantom{$\circ$} \phantom{$\circ$} \phantom{$\circ$} &     \textbf{0.02$\pm$0.02} &     \textbf{0.02$\pm$0.01} &     \textbf{0.02$\pm$0.02} &     \textbf{0.02$\pm$0.02} &     \textbf{0.02$\pm$0.01} &     \textbf{0.02$\pm$0.02} \\
       54 &                                                                0.19$\pm$0.03 $\circ$ $\circ$ $\circ$ \phantom{$\circ$} $\circ$ $\circ$ &     \textbf{0.16$\pm$0.03} &     \textbf{0.16$\pm$0.03} &     \textbf{0.16$\pm$0.03} &     \textbf{0.16$\pm$0.03} &              0.18$\pm$0.03 &     \textbf{0.16$\pm$0.03} \\
      181 &  \underline{0.39$\pm$0.04} \phantom{$\circ$} \phantom{$\circ$} \phantom{$\circ$} \phantom{$\circ$} \phantom{$\circ$} \phantom{$\circ$} &     \textbf{0.38$\pm$0.03} &  \underline{0.39$\pm$0.05} &  \underline{0.39$\pm$0.04} &     \textbf{0.38$\pm$0.03} &  \underline{0.39$\pm$0.04} &  \underline{0.39$\pm$0.04} \\
     1049 &  \underline{0.09$\pm$0.02} \phantom{$\circ$} \phantom{$\circ$} \phantom{$\circ$} \phantom{$\circ$} \phantom{$\circ$} \phantom{$\circ$} &  \underline{0.09$\pm$0.02} &  \underline{0.09$\pm$0.02} &     \textbf{0.08$\pm$0.01} &  \underline{0.09$\pm$0.01} &  \underline{0.09$\pm$0.01} &     \textbf{0.08$\pm$0.01} \\
\bottomrule
\end{tabular}

    }
    \caption{Stage-Wise Results in Isolated Configuration for scikit-learn.}
    \label{tab:namlresults:monotone:python}
\end{table}

\newpage
\begin{table}[h!]
    \centering
    \resizebox{\textwidth}{!}{
        \begin{tabular}{rrrrrrrr}
\toprule
 openmlid &                                                                                                            auto-weka &                                                                                                               mlplan &                  primitive &                  filtering &                       meta &                     tuning &                       full \\
\midrule
        3 &       \textbf{0.0$\pm$0.0} \phantom{$\circ$} \phantom{$\circ$} \phantom{$\circ$} \phantom{$\circ$} \phantom{$\circ$} &  \underline{0.01$\pm$0.01} \phantom{$\circ$} \phantom{$\circ$} \phantom{$\circ$} \phantom{$\circ$} \phantom{$\circ$} &       \textbf{0.0$\pm$0.0} &       \textbf{0.0$\pm$0.0} &       \textbf{0.0$\pm$0.0} &      \textbf{0.0$\pm$0.01} &       \textbf{0.0$\pm$0.0} \\
       12 &      \textbf{0.02$\pm$0.0} \phantom{$\circ$} \phantom{$\circ$} \phantom{$\circ$} \phantom{$\circ$} \phantom{$\circ$} &                        0.04$\pm$0.01 \phantom{$\circ$} \phantom{$\circ$} $\circ$ \phantom{$\circ$} \phantom{$\circ$} &  \underline{0.03$\pm$0.01} &  \underline{0.03$\pm$0.01} &     \textbf{0.02$\pm$0.01} &  \underline{0.03$\pm$0.01} &  \underline{0.03$\pm$0.01} \\
       54 &  \underline{0.18$\pm$0.02} \phantom{$\circ$} \phantom{$\circ$} \phantom{$\circ$} \phantom{$\circ$} \phantom{$\circ$} &  \underline{0.18$\pm$0.03} \phantom{$\circ$} \phantom{$\circ$} \phantom{$\circ$} \phantom{$\circ$} \phantom{$\circ$} &     \textbf{0.17$\pm$0.02} &     \textbf{0.17$\pm$0.02} &  \underline{0.18$\pm$0.02} &     \textbf{0.17$\pm$0.02} &  \underline{0.18$\pm$0.05} \\
\bottomrule
\end{tabular}
    }
    \caption{Stage-Wise Results in Monotone Configuration for WEKA.}
    \label{tab:namlresults:monotone:python}
\end{table}

\newpage
\begin{table}[h!]
    \centering
    \resizebox{\textwidth}{!}{
        \begin{tabular}{rrrrrrrr}
\toprule
 openmlid &                                                                                                                               asklearn &                  primitive &                      scale &                  filtering &                       meta &                     tuning &                       full \\
\midrule
        3 &       \textbf{0.0$\pm$0.0} \phantom{$\circ$} \phantom{$\circ$} \phantom{$\circ$} \phantom{$\circ$} \phantom{$\circ$} \phantom{$\circ$} &       \textbf{0.0$\pm$0.0} &       \textbf{0.0$\pm$0.0} &       \textbf{0.0$\pm$0.0} &                        nan &       \textbf{0.0$\pm$0.0} &       \textbf{0.0$\pm$0.0} \\
       12 &  \underline{0.03$\pm$0.02} \phantom{$\circ$} \phantom{$\circ$} \phantom{$\circ$} \phantom{$\circ$} \phantom{$\circ$} \phantom{$\circ$} &     \textbf{0.02$\pm$0.02} &     \textbf{0.02$\pm$0.01} &     \textbf{0.02$\pm$0.01} &     \textbf{0.02$\pm$0.02} &     \textbf{0.02$\pm$0.01} &  \underline{0.03$\pm$0.02} \\
       54 &                                                                          0.19$\pm$0.03 $\circ$ $\circ$ $\circ$ $\circ$ $\circ$ $\circ$ &     \textbf{0.16$\pm$0.03} &     \textbf{0.16$\pm$0.03} &     \textbf{0.16$\pm$0.03} &     \textbf{0.16$\pm$0.03} &  \underline{0.17$\pm$0.03} &  \underline{0.17$\pm$0.03} \\
      181 &  \underline{0.39$\pm$0.04} \phantom{$\circ$} \phantom{$\circ$} \phantom{$\circ$} \phantom{$\circ$} \phantom{$\circ$} \phantom{$\circ$} &  \underline{0.38$\pm$0.03} &              0.39$\pm$0.05 &  \underline{0.39$\pm$0.04} &  \underline{0.39$\pm$0.02} &  \underline{0.38$\pm$0.03} &     \textbf{0.37$\pm$0.05} \\
     1049 &  \underline{0.09$\pm$0.02} \phantom{$\circ$} \phantom{$\circ$} \phantom{$\circ$} \phantom{$\circ$} \phantom{$\circ$} \phantom{$\circ$} &  \underline{0.09$\pm$0.02} &  \underline{0.09$\pm$0.02} &     \textbf{0.08$\pm$0.02} &  \underline{0.09$\pm$0.01} &               0.1$\pm$0.02 &  \underline{0.09$\pm$0.01} \\
\bottomrule
\end{tabular}
    }
    \caption{Stage-Wise Results in Monotone Configuration for scikit-learn.}
    \label{tab:namlresults:monotone:python}
\end{table}

\newpage
\section{Wall-Times per Stage}
The following tables show the average runtime of \naml in the different stages per dataset.
\begin{table}[h!]
    \centering
    \resizebox{.7\textwidth}{!}{
        \begin{tabular}{rrrrrrrr}
\toprule
 openmlid &  probing &  scaling &  filtering &  meta &  tuning &  validation &  Total \\
\midrule
        3 &      123 &        0 &         65 &    80 &      28 &         127 &    423 \\
       12 &      558 &        0 &        547 &    33 &      58 &         168 &   1364 \\
       54 &       19 &        0 &         17 &    26 &      22 &          42 &    126 \\
      181 &       27 &        0 &          6 &    11 &      15 &          26 &     85 \\
     1049 &       62 &        0 &         44 &     8 &      21 &          41 &    176 \\
     1067 &       53 &        0 &         34 &    21 &      19 &          42 &    169 \\
     1111 &     3600 &        0 &          0 &     0 &       0 &           0 &   3600 \\
\bottomrule
\end{tabular}

    }
    \caption{Average time spent in stage per dataset in WEKA.}
    \label{tab:stagetimes:java}
\end{table}

\begin{table}[h!]
    \centering
    \resizebox{.7\textwidth}{!}{
        \begin{tabular}{rrrrrrrr}
\toprule
 openmlid &  probing &  scaling &  filtering &  meta &  tuning &  validation &  Total \\
\midrule
        3 &       31 &       25 &         49 &  1122 &     510 &          21 &   1758 \\
       12 &      202 &      772 &        544 &  2082 &       0 &           0 &   3600 \\
       54 &        9 &       49 &         58 &  1230 &     373 &           1 &   1720 \\
      181 &       18 &       75 &         91 &  2415 &     394 &          25 &   3018 \\
     1049 &        6 &       49 &         48 &  2049 &     420 &          30 &   2602 \\
     1067 &        7 &       52 &         54 &  3410 &      77 &           0 &   3600 \\
     1111 &      746 &      381 &        234 &  2239 &       0 &           0 &   3600 \\
\bottomrule
\end{tabular}
    }
    \caption{Average time spent in stage per dataset in scikit-learn.}
    \label{tab:stagetimes:java}
\end{table}

\end{document}